\documentclass{article}
\usepackage{spconf,amsmath,graphicx}
\usepackage{multirow}
\usepackage{comment}
\usepackage{url}
\usepackage{hyperref}
\usepackage{fancyhdr, lipsum}

\fancypagestyle{firstpage}{
   \fancyhf{} 
   \newcommand{\changefont}{%
    \fontsize{9}{11}\selectfont
    }
    \headheight 40pt

   \cfoot{\thepage}
   \fancyhead[L]{\changefont © 20XX IEEE.  Personal use of this material is permitted.  Permission from IEEE must be obtained for all other uses, in any current or future media, including reprinting/republishing this material for advertising or promotional purposes, creating new collective works, for resale or redistribution to servers or lists, or reuse of any copyrighted component of this work in other works.}
}
\title{Parameter Blending for Multi-Camera Harmonization for Automotive Surround View Systems}
\name{Yuzhuo Ren, Yining Deng, David Pajak, Robin Jenkin, Niranjan Avadhanam, Varsha Hedau}
\address{NVIDIA, Santa Clara, CA, USA}
\begin{document}
%
\maketitle

\thispagestyle{firstpage}
\begin{abstract}
In a surround view system, the image color and tone captured by multiple cameras can be different due to cameras applying auto white balance (AWB), global tone mapping (GTM) individually for each camera. The color and brightness along stitched seam location may look discontinuous among multiple cameras which impacts overall stitched image visual quality. To improve the color transition between adjacent cameras in stitching algorithm, we propose harmonization algorithm which applies before stitching to adjust multiple cameras' color and tone so that stitched image has smoother color and tone transition between adjacent cameras. Our proposed harmonization algorithm consists of AWB harmonization and GTM harmonization leveraging Image Signal Processor (ISP)'s AWB and GTM metadata statistics. Experiment result shows that our proposed algorithm outperforms global color transfer method in both visual quality and computational cost.
\end{abstract}
\begin{keywords}
surround view system, color harmonization, auto white balance, global tone mapping
\end{keywords}
\section{Introduction}
\label{sec:intro}

Surround view system is to composite images captured from multiple cameras with overlapped field of view to a wider field of view image. Surround view system has various applications in virtual reality, robotics, autonomous driving, etc. Surround view algorithm relies on camera calibration \cite{hedi2012system, ren2021camera} and feature matching to align images and then optimize seam location (Graph Cut \cite{li2016optimal, zhang2019seamless}, seam carving \cite{avidan2007seam}, dynamic seam placement \cite{ren2023image}, etc) and apply blending algorithm \cite{bellavia2017dissecting, niu2018color, xu2021color} to reduce seam visibility. Surround view quality assessment evaluates geometric distortion which caused by misalignment and photometric distortion \cite{madhusudana2019subjective, qureshi2012quantitative, ren2022image, szeliski2007image, ren2023stitching}. Photometric distortion is caused by color difference from multiple cameras. Our proposed harmonization algorithm is to reduce photometric distortion in surround view system. The color difference among different cameras is more obvious especially when different cameras are in different lighting conditions. For example, while ego car driving inside garage, the lightening conditions of inside and outside garage are very different at night which makes surround view cameras under different lightening. Applying a stitching algorithm using unharmonized camera images as input results in visible color difference and stitching seams which decreases image visual quality and user experience. 

\begin{figure}
\centering
\includegraphics[height=2.8cm]{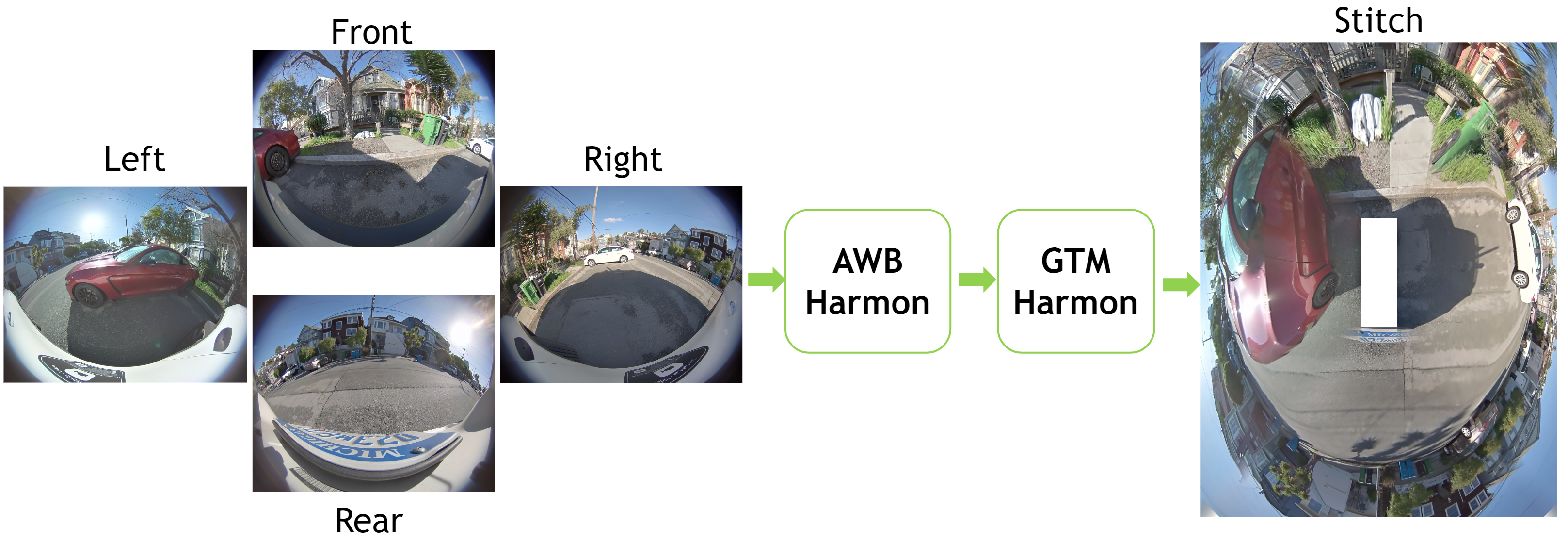}
\caption{Input to our harmonization algorithm are camera image and its metadata (i.e., auto white balance (AWB) and global tone mapping (GTM)). Our algorithm first conduct AWB harmonization to harmonize color and then followed by GTM harmonization to harmonize luminance. The harmonized images are then used for surround view stitching.}
\label{overview}
\end{figure}

There are researches harmonizing color within single camera image to make foreground and background color coherent \cite{cohen2006color, cong2022high, guerreiro2023pct, mertens2007exposure}. Existing solution for harmonize multi-images selects one camera as reference, and then apply global color transfer \cite{reinhard2001color} to match source camera’s global statistics (i.e., mean, standard deviation) to that of reference image. There are several drawbacks of the existing solution: 1) It is sensitive to reference camera selection, i.e., different reference camera generates different stitched image; 2) It is not able to handle the case when different cameras see different scenes due to disparity; 3) It usually involves selecting the most representative color patch for color statistic computation, such as the overlapping ground region where both cameras can see the same content, or local segments matching between the cameras, which are computational intensive and may not be able to fit into real time surround view system.

We propose harmonization algorithm to harmonize four fisheye cameras mounted on left mirror, front (hood), right mirror, rear (trunk) of vehicle to deliver better automotive in-cabin user experience. Our proposed multi-camera harmonization algorithm leverages Image Signal Processor (ISP) statistics, i.e., auto white balance (AWB) \cite{domabi2022weighted, afifi2022auto} and global tone mapping (GTM) \cite{deng2021enhanced} to gradually blend image's color and luminance from image center to image boundary, where stitched seam locates. In image center, color and luminance maintain the same as original image, from image center to image boundary, AWB and GTM are gradually blended with its adjacent cameras' AWB and GTM to achieve smooth transition between cameras. Our proposed method has several advantages: 1) Our method blends AWB and GTM parameters between adjacent cameras which avoids explicitly selecting any camera as reference; 2) AWB and GTM is more accurate than color statistics from overlap patch especially when overlap patch is not able to represent image color; 3) directly using AWB and GTM from ISP avoids computing image color statistics from overlap patches which saves a lot of computes.

\section{Methodology}
\label{sec:methodology}
We propose color harmonization algorithm which leverages camera metadata information to avoid color statistics calculation from image. We uniquely designed metadata in ISP for harmonization feature to achieve better stitching quality and to fit into compute budget to enable real time system. Our proposed metadata includes: 1) auto white balance gain (AWB) in RGB channel respectively, which is a 3x1 vector and 2) 256 point global tone mapping (GTM) look up table, same for RGB channel. Our system consists AWB harmonization module followed by GTM harmonization module applied to every adjacent camera pairs, as shown in Fig.~\ref{overview}, to provide better end to end color harmonization solution for stitching. 
\subsection{Blending Curve}
In order to achieve smoother color transition between cameras, standard logistic function is applied to harmonize AWB and GTM parameters, which is in Eq.\ref{eq:logistic_standard} 
\begin{equation}
\begin{split}
f_{\text{standard}}(x)=1/(1+\exp(-x)), x\in[-6, 6],
\end{split}
\label{eq:logistic_standard}
\end{equation}
where $x$ is bounded within [-6, 6] because we want $f_{\text{standard}}(x)$ bounded to (0, 1) \footnote[1]{$f_{\text{standard}}(-6) = 0.0025$, $f_{\text{standard}}(6) = 0.9975$}. Logistic function is smoother than linear function in terms of transition. For example, if left side and right side of the image are harmonized to different color/brightness direction, i.e., image left side brightness needs to be decreased(increased) after harmonization while right side brightness needs to be increased(decreased) after harmonization, logistic curve provides smoother transition in the middle of image than linear curve. In order to make the logistic curve $x$ index starts from 0, and $f(x)$ ranges $[0, 1]$, our modified logistic function $f(x)$ is defined as following:
\begin{equation}
\begin{split}
f(x)=\text{SCALE}/(1+\exp(-x + 6)) + \text{SHIFT}, x\in[0, 12],
\end{split}
\label{eq:logistic}
\end{equation}
where $\text{SCALE} =1.005$, $\text{SHIFT}=-0.0025$. Our modified logistic curve is scaled up a bit and shifted down a bit to make the curve $f(0) = 0 - \delta$, $f(12) = 1 + \delta$, $\delta$ is a very small positive number. Then $f(x)$ is bounded to [0, 1] by truncating any values beyond [0, 1].

\begin{figure}
\centering
\includegraphics[height=5.8cm]{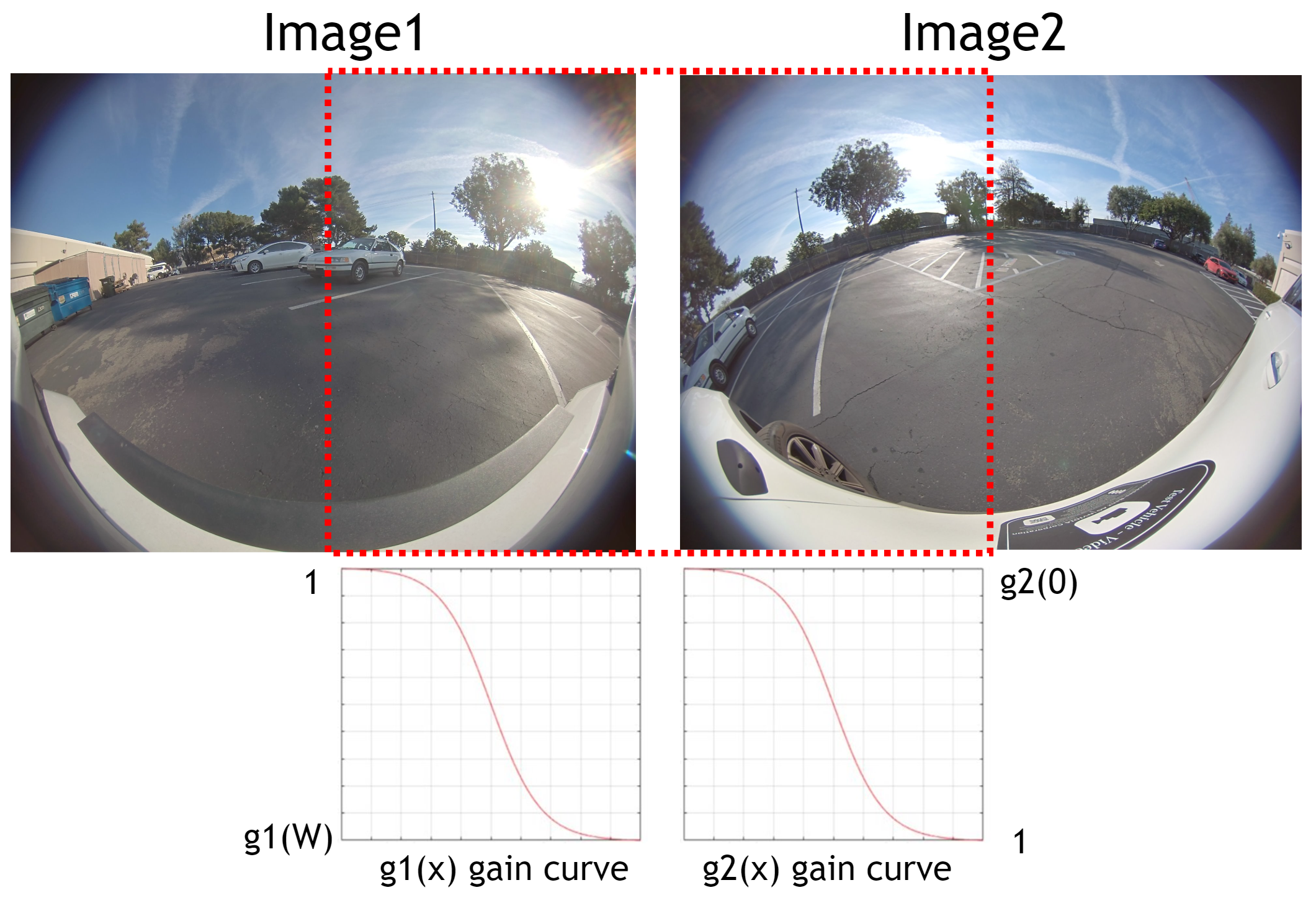}
\caption {Image1 is brighter along overlap boundary thus it has a decreased gain curve from middle column to boundary; Image 2 is darker along overlap boundary thus it has an increased gain curve from middle column to boundary.}
\label{fig:gain}
\end{figure}

Assume right side of camera 1 overlaps with left side of camera 2, $g_1 (x)$ and $g_2 (x)$ are the AWB or GTM gain curve for right side of camera 1 and left side of camera 2 respectively, as shown in Fig.\ref{fig:gain}. In the middle column of image, pixel values maintain the same to reflect original image's color and tone, thus AWB gain and GTM gain is 1, i.e., 
\begin{equation}
g_1 (\text{W}/2) = g_2 (\text{W}/2) = 1,
\end{equation}
\begin{equation}
\begin{split}
\label{eq:gx}
g_1 (x) &= 1 + (g_1 (\text{W}-1) – 1) \cdot f(\frac{x-\text{W}/2}{\text{W}/2} \cdot 12), \\
\text{where}, x& \in [\text{W}/2, \text{W} - 1], \\
g_2 (x) &= g_2 (0) + (1 - g_2 (0)) \cdot f(\frac{x}{\text{W}/2} \cdot 12), \\
\text{where}, x&\in [0, \text{W}/2),
\end{split}
\end{equation}
where W denotes image width, $f(x)$ \footnote[2]{image column index $x$ in $f(x)$ in Eq.\ref{eq:gx} is linearly shifted and interpolated to $[0, 12]$ to be able to apply Eq.\ref{eq:logistic}} is defined in Eq.\ref{eq:logistic}. In the boundary of image (i.e., $x= \text{W}-1$ and $x=0$), AWB gain and GTM gain is computed based on averaged AWB and GTM from two cameras. We will define $g_1(\text{W}-1)$ and $g_2(0)$ in next two sub sessions. From middle column of image to the image boundary, AWB gain and GTM gain is interpolated following the logistic curve $g(x)$.
\subsection{AWB Harmonization}
AWB gain in image boundary column is defined as:
\begin{equation}
\label{eq:awb}
\begin{split}
g_1^{\text{AWB}} (\text{W} -1) &= ((\text{AWB}_1 + \text{AWB}_2 ) / 2) / \text{AWB}_1, \\
g_2^{\text{AWB}} (0) &= ((\text{AWB}_1 + \text{AWB}_2 ) / 2) / \text{AWB}_2,
\end{split}
\end{equation}

where $\text{AWB}_1$ and $\text{AWB}_2$ are camera 1 and camera 2’s AWB value from ISP. 

\subsection{GTM Harmonization}
\begin{figure}
\centering
\includegraphics[height=3.3cm]{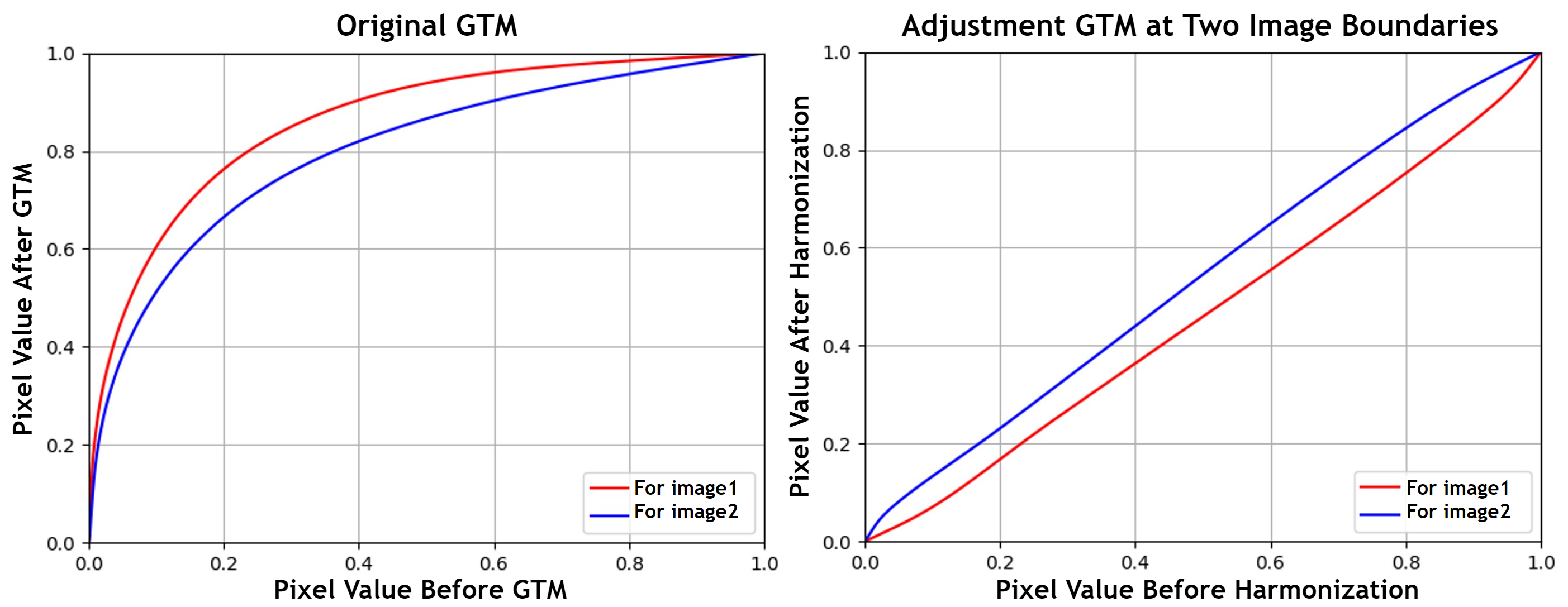}
\caption{Left: original GTM look up table from metadata for two adjacent camera image. Image1 has brighter GTM curve than image2 before harmonization. Right: Harmonized GTM look up table at two image boundary. Image1's brightness is reduced while image2's brightness is increased after harmonization.}
\label{GTM}
\end{figure}

GTM gain in image boundary column is defined as:
\begin{equation}
\label{eq:gtm}
\begin{split}
g_1^{\text{GTM}} (\text{W} -1) &= \text{LUT}_1^{\text{GTM}} (\text{pixelVal}_1 ) / \text{pixelVal}_1, \\
g_2^{\text{GTM}} (0) &= \text{LUT}_2^{\text{GTM}} (\text{pixelVal}_2 ) / \text{pixelVal}_2,
\end{split}
\end{equation}
where $\text{LUT}_1$ and $\text{LUT}_2$ are harmonized GTM 256 point look up table (LUT). $\text{LUT}_1^{\text{GTM}}$ and $\text{LUT}_2^{\text{GTM}}$ are obtained by interpolation between original and averaged GTM LUT. Fig.~\ref{GTM} shows example of GTM before and after harmonization.
The adjusted pixel value after AWB and GTM harmonization is the original pixel value multiplied by $g^{\text{AWB}} \cdot g^{\text{GTM}}$.

\begin{figure*}
\centering
\includegraphics[width=16cm]{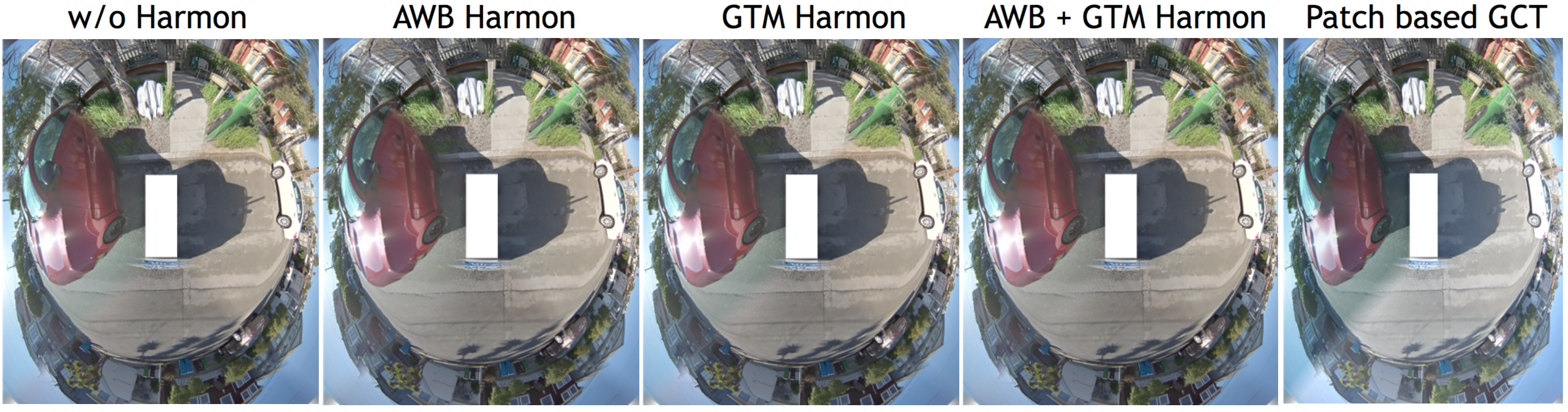}
\caption{Ablation on AWB and GTM Harmonization. AWB harmonization reduces greenish artifact on the left camera. GTM harmonization reduces luminance difference between right and rear camera.}
\label{fig:result}
\end{figure*}

\begin{figure}
\centering
\includegraphics[height=3cm]{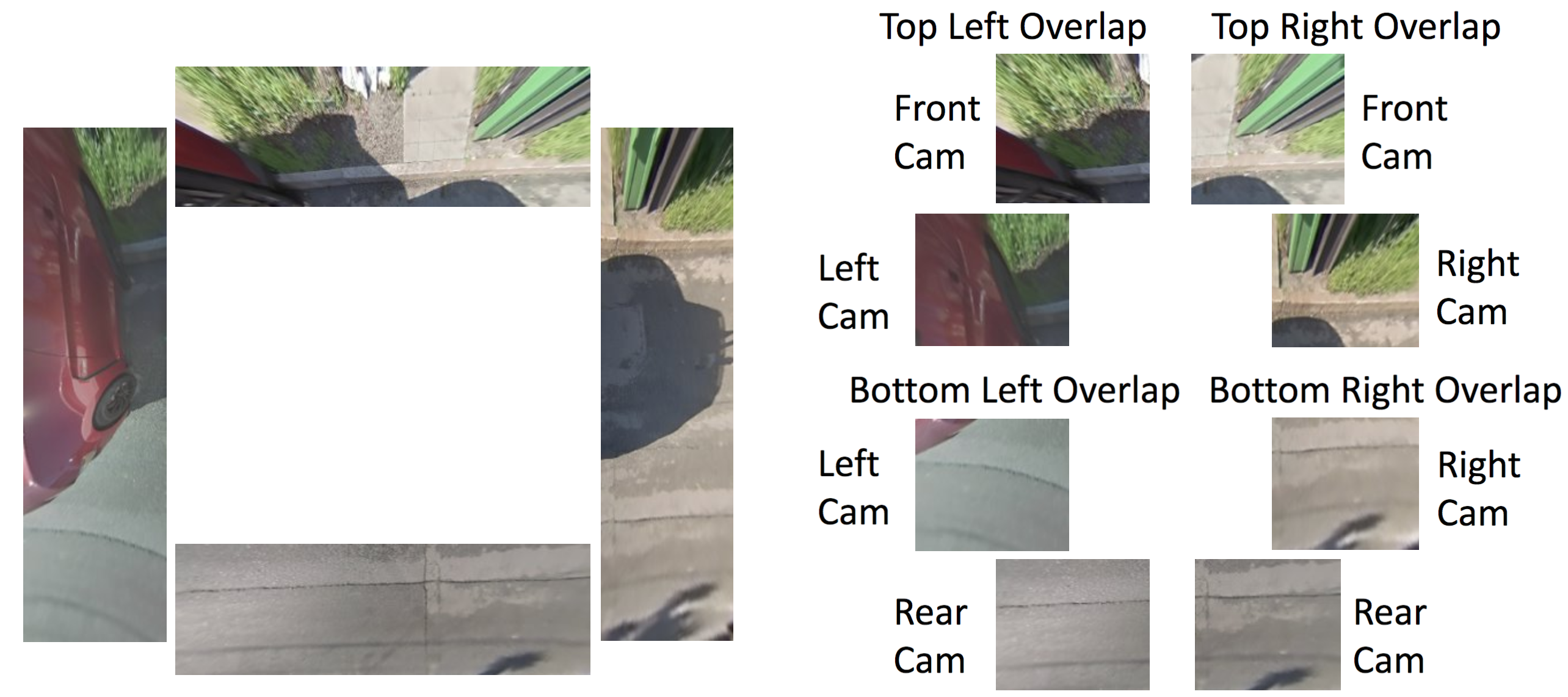}
\caption{Ground projection and overlaps. Different cameras may not see the same objects in the overlap region.}
\label{fig:overlap}
\end{figure}

\begin{figure}
\centering
\includegraphics[height=11cm]{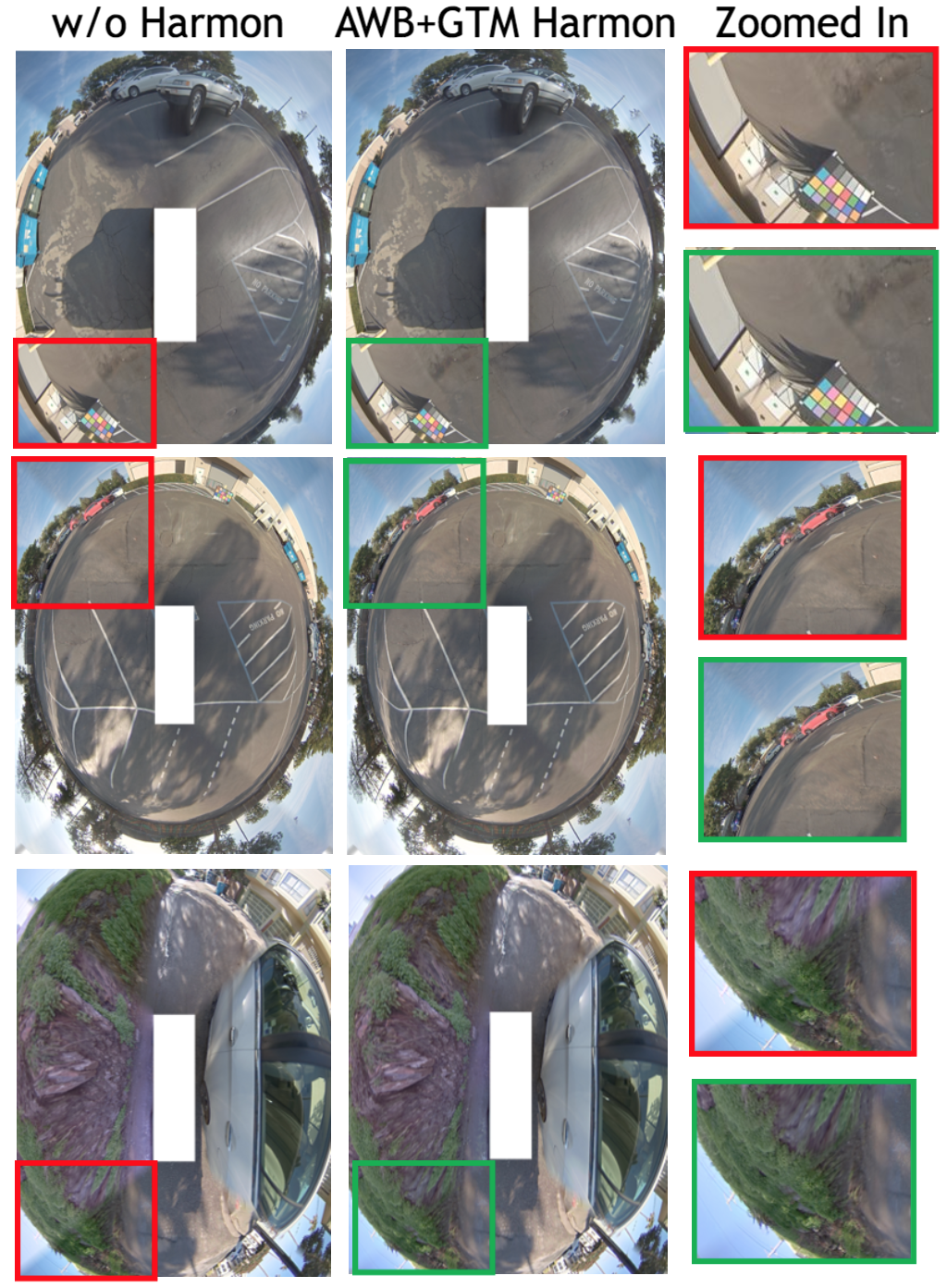}
\caption{Visual result comparison.}
\label{fig:result_more}
\end{figure}

\section{Experiments}
 
We evaluate our proposed harmonization algorithm in various aspects. We compare our method with patch based GCT \cite{reinhard2001color} method in both subjective quality and runtime.

\subsection{Evaluation Dataset}
We are not aware of existing datasets which contain synchronized
multi-camera images with ISP AWB and GTM parameters to evaluate our proposed method. So we collected our evaluation dataset. Our dataset contains four fisheye cameras's images (200 FOV RGB sensor at 30 FPS with resolution 1920x1536) mounted on left, front, right, rear of vehicle with fixed exposure and its ISP AWB and GTM data. Our dataset contains a wide varieties of scenarios in total 7 sessions, including daytime, sunset, nighttime garage, nighttime highway, nighttime traffic lights, green grass and red car. Each session is a 10 second video. 

\subsection{Implementation Details}
\subsubsection{Patch based Global Color Transfer (GCT)} Four fisheye camera images are projected onto the ground using perspective projection \cite{szeliski2007image} based on camera calibration parameters. Then ground overlap region between adjacent cameras is computed. Color statistics (mean and std) of overlapping patch is computed which is used as the whole image color statistics. Then color transfer equation in \cite{reinhard2001color} is applied to transfer global color statistics to match reference image. Left and right camera uses front camera as reference; Rear camera uses right camera as reference. We found that computing color statistics from overlap region rather than from entire image gives better color transfer result because various objects exists in different cameras which makes whole image color statistics not representative for the true brightness of the image. 

\subsubsection{Surround View Stitching} We generate surround view stitching image result to have a better visualization of before harmonization and after harmonization effect. In the stitched image, we can clearly see the color and brightness transition from one camera to another camera. We use camera intrinsic and extrinsic calibration parameters to project cameras onto a bowl top projection to generate stitched image from a top down perspective. Within an elliptical radius around ego vehicle, bowl top projection is equivalent to top down perspective projection; Beyond the elliptical radius, height of the bowl wall increases following the parabola curve. For details regarding our surround view stitching implementation, we recommend the readers to refer to this documentation \cite{surroudview}.   
\subsection{Ablation on AWB and GTM Harmonization}
To validate the effectiveness of both AWB and GTM harmonization, we show harmonization result with only AWB harmonization and with only GTM harmonization respectively in Fig.\ref{fig:result}. AWB harmonization reduces the color difference between left camera and rear camera, i.e., the greenish artifact on the ground is significantly reduced. The greenish artifact is caused by the challenge of AWB algorithm in this case. GTM harmonization reduces brightness difference between right camera and rear camera. Applying both AWB harmonization and GTM harmonization can reduce both color difference and luminance difference. GCT method suffers from reference image selection and representativeness of ground overlap patch. Left camera is not well harmonized with rear camera because rear camera is using right camera as color transfer reference. Left camera is not well harmonized with front camera because left and front cameras see different objects in overlap region due to disparity which makes it sensitive to patch selection, as shown in Fig.~\ref{fig:overlap}.
\subsection{Subjective Quality Evaluation}
We invited 10 subjects to visually evaluate stitched image for color and luminance transition smoothness between adjacent cameras. We show stitched video result side by side comparing without applying harmonization versus patch based GCT versus our metadata based harmonization method. All subjects agree that our proposed harmonization method significantly improves the color and luminance transition among cameras and results in a better visual quality and better than patch based GCT methods. Fig.~\ref{fig:result_more} shows more visual results.

\subsection{Runtime Evaluation}
We implemented patch based GCT method and our proposed metadata approach in CUDA and evaluated runtime on NVIDIA DRIVE Orin SoC, which are shown in Table \ref{table:runtime}. Our metadata approach saves 50\% computation cost compared to patch based GCT method.

\begin{table}[]
\begin{tabular}{|l|lr|c|}
\hline
                                                                                               & \multicolumn{2}{l|}{Runtime Breakdown (ms)}                                                                                      & \multicolumn{1}{l|}{Total (ms)} \\ \hline
\multirow{3}{*}{\begin{tabular}[c]{@{}l@{}}Patch based\\ Global Color\\ Transfer\end{tabular}} & \multicolumn{1}{l|}{project to ground}                                                                       & 0.32 & \multirow{3}{*}{0.52}      \\ \cline{2-3}
                                                                                               & \multicolumn{1}{l|}{\begin{tabular}[c]{@{}l@{}}compute color statistics\\ for the ground patch\end{tabular}} & 0.1  &                            \\ \cline{2-3}
                                                                                               & \multicolumn{1}{l|}{apply global transfer}                                                                   & 0.1  &                            \\ \hline
Ours                                                                                           & \multicolumn{1}{l|}{apply logistic curve}                                                                    & 0.26 & 0.26                       \\ \hline
\end{tabular}
\caption{Runtime comparison of patch based global color transfer method and our method.}
\label{table:runtime}
\end{table}
\section{Conclusion}
\label{sec:conclusion}
We proposed parameter blending for multi-camera harmonization for automotive surround view system. We proposed applying logistic curve to blend AWB and GTM parameters to achieve smoother color transition among adjacent cameras. We compared our solution to patch based global color transfer method. The experiment results demonstrated that metadata based logistic blending method outperforms in both visual quality and runtime.

\newpage
\bibliographystyle{IEEEbib}
\bibliography{egbib}


\end{document}